# A Slice Sampler for Restricted Hierarchical Beta Process with Applications to Shared Subspace Learning

**Sunil Kumar Gupta**  **Dinh Phung**  **Svetha Venkatesh**

Center for Pattern Recognition and Data Analytics (PRaDA)

Deakin University, Australia

{sunil.gupta, dinh.phung, svetha.venkatesh}@deakin.edu.au

## Abstract

Hierarchical beta process has found interesting applications in recent years. In this paper we present a modified hierarchical beta process prior with applications to hierarchical modeling of multiple data sources. The novel use of the prior over a hierarchical factor model allows factors to be shared across different sources. We derive a slice sampler for this model, enabling tractable inference even when the likelihood and the prior over parameters are non-conjugate. This allows the application of the model in much wider contexts without restrictions. We present two different data generative models – a linear Gaussian-Gaussian model for real valued data and a linear Poisson-gamma model for count data. Encouraging transfer learning results are shown for two real world applications – text modeling and content based image retrieval.

## 1 Introduction

Hierarchical modeling is becoming increasingly popular in Bayesian statistics – it allows joint modeling of multiple data sources, permitting shared patterns. A successful example of hierarchy in text modeling is the hierarchical Dirichlet process (HDP) [18]. HDP is used as a prior over a mixture model allowing shared mixture components (topics) across multiple data sources. Each data point from a data source is modeled using 1-out-of-$K$ topics, shared across different sources. In many applications, where it may not be easy to separate out data into $K$ mutually exclusive groups, an appropriate approach is factor analysis, where each data point is modeled using $P$-out-of-$K$ factors. The benefits of joint modeling can still be retained by sharing these factors across different sources.

Previous works on nonparametric factor analysis have used beta processes. One of the earliest work develops the Indian buffet process (IBP) and uses it as a nonparametric prior over infinite binary matrices [6]. Similar works differing in inference methods are proposed in [15, 14, 5]. A significant contribution by Thibaux and Jordan [19] uncovers that the IBP is a predictive process for an underlying beta process with concentration parameter equaling one. In addition, they propose a hierarchical beta process (HBP) model and demonstrate its use for document classification in a *single* data source. The use of HBP can be extended to modeling multiple data sources by utilizing it as a nonparametric prior for *factor* models, in a similar way to HDP being used as a nonparametric prior for *mixture* models. However, when combining HBP with the factor model, inference is often intractable. This is mainly due to the need to integrate out the parameters when invoking new factors according to the HBP prior. For efficient computation of this integration, conjugacy is required between the parameter priors and the likelihood. However, the requirement of the conjugacy restricts the applicability of this model. One approach to deal with the non-conjugacy issues is through explicit representation of all the samples drawn from beta process. This requires representation of an infinite set of atoms - a practically impossible task. To circumvent this problem, one possible solution is to truncate the beta process. However, a pre-specified truncation level is arbitrary. To avoid this arbitrary truncation for a IBP based factor model, Teh et al [17] use a slice sampling technique, which turns the infinite representation problem into a finite representation problem. A similar approach for Dirichlet processes was taken by Walker [20]. However, a slice sampler for HBP based factor model is *yet* to be developed.

An attempt on nonparametric, hierarchical factor analysis using HBP was made in [8]. However, inference for this model has many *approximation* steps (refer section 3.1.2 in [8]), which risk sampling from incorrect posterior distributions. Thus *more accurate inference* methods *need to be developed*.

In this paper, we take the HBP prior and replace the upper beta process layer by an IBP prior, i.e. the concentration parameter of the underlying beta process is fixed to one. We refer to this modified hierarchical prior as the

*Restricted Hierarchical Beta Process* (R-HBP). This modification allows us to use the stick breaking construction of IBP [17] and explicitly represent the samples of the underlying beta process. To handle the resultant infinite representation problem, *we present a novel slice sampler* for the proposed R-HBP, deriving an infinite limit, necessary for tractable posterior over stick-weights of IBP. At the data source level, we still retain beta processes with arbitrary concentration parameters, allowing arbitrary sharing between data sources.

For the hierarchical factor analysis (i.e. factor analysis using R-HBP prior), we propose two different data models – a linear Gaussian data model for real-valued data and a Poisson-gamma model for count data. For the linear Poisson-gamma model, we present novel inference using auxiliary variables. Through synthetic experiments, we illustrate the behavior of our model, demonstrating that it correctly discovers all parameters including the dimensionalities of the factor subspaces. We further demonstrate the proposed models for two real world tasks – text modeling and content based image retrieval. For text modeling, which is demonstrated on NIPS 0-12 dataset (constructed from 12 years of NIPS proceedings), we successfully show the benefits of transfer learning. In addition, we show that linear Poisson-gamma model achieves much better perplexity than other models such as linear Gaussian model or HDP mixture model. For content based image retrieval, using NUS-WIDE animal dataset, we demonstrate that our method outperforms recent state-of-the-art methods.

Our main contributions are:

- A slice sampler for the proposed R-HBP model – we derive an infinite limit, which is necessary for tractable posterior over stick-weights of IBP.

- A novel, auxiliary variable, Gibbs sampler for the linear Poisson-gamma model.

- Demonstration of the proposed models for text modeling and content based image retrieval.

The rest of this paper is organized as follows: Section 2 presents brief background on the related nonparametric priors. Section 3 describes the modeling distributions for nonparametric hierarchical factor analysis. Section 4 presents the inference covering the novel slice sampling along with Gibbs sampling. Section 5 demonstrates the experiments using synthetic and real-world text and image data. Section 6 concludes this paper.

## 2 Preliminaries

### 2.1 Hierarchical Modeling using Beta Process

Let $\Omega$ be the set of all outcomes and $\mathscr{F}$ be its sigma algebra. We denote a beta process [10] as $B \sim \mathrm{BP}(\gamma_0, B_0)$ where $\gamma_0$ is a positive concentration parameter and $B_0$ is a base measure on $\Omega$. The beta process is a completely random measure [13], implying that if $C_1, \ldots, C_n$ are the disjoint sets in $\Omega$, the measures $B(C_1), \ldots, B(C_n)$ are independent random variables and the draws from a beta process are discrete with probability one. Using this property, a draw $B$ from the beta process $\mathrm{BP}(\gamma_0, B_0)$ can be written as $B = \sum_k \beta_k \delta_{\phi_k}$ where $\phi_k \in \Omega$ is an atom drawn from $B_0$ and $\beta_k$ is a random weight assigned to $\phi_k$ according to the following

$$\beta_k \sim \mathrm{beta}(\gamma_0 B_0(\phi_k), \gamma_0(1 - B_0(\phi_k))), k = 1, 2, \ldots \quad (1)$$

If $B_0 = \sum_k \xi_k \delta_{\phi_k}$ (i.e. discrete), $B_0(\phi_k) = \xi_k$. For a continuous $B_0$, $\{(\phi_k, \beta_k)\}$ are i.i.d. draws from a Poisson process on the product space $\Omega \times [0, 1]$ with a Lévy measure [12].

Building a hierarchy over beta processes, Thibaux and Jordan [19] introduced a hierarchical beta process (HBP) prior that allows the sharing of the atoms (drawn from a beta process) across multiple data sources. A hierarchical beta process is constructed in the following way : $B \sim \mathrm{BP}(\gamma_0, B_0)$ is a draw from a beta process with base measure $B_0$ and there emanate $J$ child beta processes using $B$ as their base measure (drawn as $A_j \sim \mathrm{BP}(\alpha_j, B)$, $j = 1, \ldots, J$). Finally, each $A_j$ is used to parametrize a Bernoulli process denoted as $\mathbf{Z}_j^{i,:} \mid A_j \sim \mathrm{BeP}(A_j)$ where $\mathbf{Z}_j^{i,:}$ denotes $i$-th row of a binary matrix $\mathbf{Z}_j$. The support of random measure $B$ is the same as that of base measure $B_0$ and is passed on to each $A_j$ (for each $j$) allowing sharing of atoms across $J$-sources.

### 2.2 Indian Buffet Process and Stick-Breaking Construction

Indian buffet process (IBP) [6] is a nonparametric predictive prior developed as a generative model for infinite binary matrices. A key property of IBP is that it is exchangeable, i.e. if $\mathbf{Z} = [z_1, \ldots, z_N]$ (where $z_i$ is a binary-valued infinite vector) then $p(z_1, \ldots, z_N) = p(z_{\sigma(1)}, \ldots, z_{\sigma(N)})$ where $\sigma$ is a permutation over $\{1, \cdots, N\}$. Given this exchangeability property, it is interesting to find out the underlying de Finetti stochastic process [4]. Investigating this question, Thibaux and Jordan [19] discovered that this underlying stochastic process is a beta process with concentration parameter equal to one, (i.e. $\gamma_0 = 1$ in Eq (1)). In most of the models, to construct the binary matrix $\mathbf{Z}$ from IBP, a sequential process [6] can be followed. However, this keeps the underlying beta process hidden. There are many applications where it is required to represent the underlying beta process explicitly. In such applications, one needs to know how to sample from the beta process underlying IBP. A solution to this problem was proposed by Teh et al [17] who derived stick-breaking construction for IBP and related this scheme to a similar stick breaking construction for Dirichlet processes.

Let us denote the beta process underlying IBP by $B \sim \mathrm{BP}(1, B_0)$ and assume that $\tau_0 = B_0(\Omega)$. Using an infi-nite set of atoms, we can write $B = \sum_l \beta_l \delta_{\phi_l}$. Now let $\{\beta_{(1)}, \beta_{(2)}, \ldots, \beta_{(K)}\}$ be the decreasing ordered representation of $\{\beta_1, \beta_2, \ldots, \beta_K\}$, where $\beta_l \sim \text{beta}\left(\frac{\tau_0}{K}, 1\right)$. Teh et al [17] showed that in the limit $K \to \infty$, the $\beta_{(k)}$ variables can be obtained from the following procedure

$$v_{(t)} \sim \text{beta}(\tau_0, 1), \quad \beta_{(k)} = v_{(k)} \beta_{(k-1)} = \Pi_{t=1}^{k} v_{(t)} \quad (2)$$

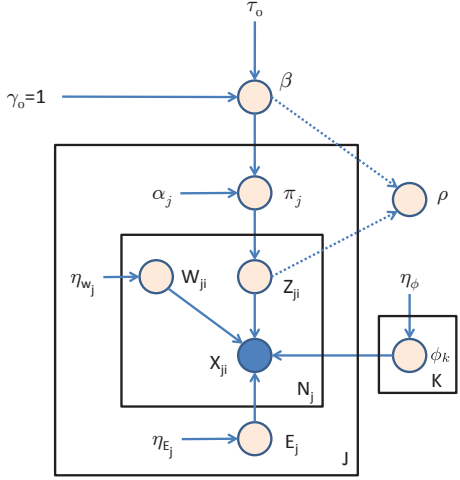

Figure 1: Directed graphical representation for the proposed nonparametric hierarchical factor model. The hyperparameters over $\mathbf{W}_j$, $\phi_k$ and $\mathbf{E}_j$ are denoted by $\eta_{w_j}, \eta_\phi$ and $\eta_{E_j}$ respectively.

We shall denote the stick-breaking construction of Eq (2) as $\beta_{(k)} \sim \text{StickIBP}(\tau_0)$. A key property of this construction is that there exists a Markov relation [17] among $\{\beta_{(1)}, \beta_{(2)}, \ldots, \beta_{(K)}\}$, i.e. $\beta_{(k)}$ is conditionally independent of $\beta_{(1:k-2)}$ given $\beta_{(k-1)}$. Formally,

$$p\left(\beta_{(k)} \mid \beta_{(1:k-1)}\right) = \tau_0 \beta_{(k-1)}^{-\tau_0} \beta_{(k)}^{\tau_0 - 1} \mathbf{1}\left(0 \leq \beta_{(k)} \leq \beta_{(k-1)}\right) \quad (3)$$

where $\mathbf{1}(Q)$ equals to one if $Q$ is true and zero otherwise. Given top $k$ variables $\{\beta_{(1)}, \beta_{(2)}, \ldots, \beta_{(k)}\}$, the other variables $\beta_l$ have the following density function [17]

$$p\left(\beta_l \mid \beta_{(1:k)}\right) = \frac{\tau_0}{K} \beta_{(k)}^{-\left(\frac{\tau_0}{K}\right)} \beta_l^{\left(\frac{\tau_0}{K}\right)-1} \mathbf{1}\left(0 \leq \beta_l \leq \beta_{(k)}\right) \quad (4)$$

### 2.3 Restricted Hierarchical Beta Process

Restricted hierarchical beta process (R-HBP) is defined as a hierarchical beta process where the parent beta process has a concentration parameter equal to one. In other words, the parent beta process in a R-HBP is a beta process for which the predictive process is an Indian buffet process. Formally,

$$B \sim \text{BP}(1, B_0), \quad A_j \sim \text{BP}(\alpha_j, B), \quad \mathbf{Z}_j^{i,:} \mid A_j \sim \text{BeP}(A_j)$$

Alternatively, using Eqs (1-2), we can write the above as

$$\beta_{(k)} \sim \text{StickIBP}(\tau_0) \quad (5)$$

$$\pi_{jk} \sim \text{beta}\left(\alpha_j \beta_{(k)}, \alpha_j (1 - \beta_{(k)})\right), \mathbf{Z}_j^{i,:} \sim \text{BeP}(\pi_j) \quad (6)$$

## 3 Hierarchical Factor Analysis using R-HBP

We now consider joint factor analysis [7, 8] and use the R-HBP prior to infer the dimensionalities of factor subspaces. Our goal is to model *multiple* data matrices $\mathbf{X}_1, \ldots, \mathbf{X}_J$ jointly using a factor matrix $\Phi = [\phi_1, \ldots, \phi_K]$ where $\phi_k \in \mathbb{R}^M$ denotes the $k$-th factor. Some of the factors in $\Phi$ may be *shared* amongst the various data sources whereas others factors are specific to *individual* sources. For each $j = 1, \ldots, J$, $\mathbf{X}_j \in \mathbb{R}^{M \times N_j}$ has a representation $\mathbf{H}_j \in \mathbb{R}^{N_j \times K}$ in the subspace spanned by $\Phi$ along with the factorization error $\mathbf{E}_j$. Formally,

$$\Pi : \begin{cases} \mathbf{X}_1 = & \Phi \mathbf{H}_1^\mathsf{T} + \mathbf{E}_1 \\ \cdots & \cdots \\ \mathbf{X}_J = & \Phi \mathbf{H}_J^\mathsf{T} + \mathbf{E}_J \end{cases} \quad (7)$$

We allow the number of factors ($K$) to grow infinitely when more data is observed. To infer the value of $K$, we represent the matrix $\mathbf{H}_j$ as an element-wise multiplication of two matrices $\mathbf{Z}_j$ and $\mathbf{W}_j$, i.e. $\mathbf{H}_j = \mathbf{Z}_j \odot \mathbf{W}_j$, where $\mathbf{Z}_j^{ik} = 1$ implies the presence of factor $\phi_k$ for $i$-th data point $\mathbf{X}_j^{:,i}$ from source $j$ and $\mathbf{W}_j^{ik}$ represents the corresponding coefficient or weight of the factor $\phi_k$. The collection of matrices $\mathbf{Z}_1, \ldots, \mathbf{Z}_J$ are now modeled using R-HBP prior. In particular, we model $i$-th row of $\mathbf{Z}_j$ as a draw from a Bernoulli process parametrized by a beta process $A_j$, i.e. $\mathbf{Z}_j^{i,:} \sim \text{BeP}(\pi_j)$ where $\pi_j \triangleq [\pi_{j1}, \ldots, \pi_{jK}]$ and $\pi_{jk}$ is defined in Eq (6). We note that sampling $\pi_{jk}$ according to Eq (6) allows sharing of the factors $\phi_k$'s across data sources. This is because $\beta_{(k)}$ (the stick-breaking weights of $B$, i.e. $\beta_{(k)} \sim \text{StickIBP}(\tau_0)$) are used for all data sources. For the factor analysis parameters, we propose two different models : a Gaussian-Gaussian model (GGM) for real-valued data and a Poisson-gamma model (PGM) for count data. The whole model can be summarized as

$$\text{PGM} : \begin{cases} \Phi^{ik} \sim \text{gamma}(a_\phi, b_\phi) \\ \mathbf{W}_j^{i,:} \sim \Pi_{k=1}^{K} \text{gamma}(a_{w_j}, b_{w_j}) \\ \mathbf{X}_j^{:,i} \mid \Phi, \mathbf{Z}_j^{i,:}, \mathbf{W}_j^{i,:} \\ \sim \text{Poisson}\left(\Phi\left(\mathbf{Z}_j^{i,:} \odot \mathbf{W}_j^{i,:}\right)^\mathsf{T} + \lambda_j\right) \end{cases} \quad (8)$$

$$\text{GGM} : \begin{cases} \Phi^{ik} \sim \mathcal{N}\left(0, \sigma_\phi^2\right), \mathbf{W}_j^{i,:} \sim \mathcal{N}\left(\mathbf{0}, \sigma_{w_j}^2 \mathbf{I}\right) \\ \mathbf{X}_j^{:,i} \mid \Phi, \mathbf{Z}_j^{i,:}, \mathbf{W}_j^{i,:} \sim \mathcal{N}\left(\Phi\left(\mathbf{Z}_j^{i,:} \odot \mathbf{W}_j^{i,:}\right)^\mathsf{T}, \sigma_{n_j}^2 \mathbf{I}\right) \end{cases} \quad (9)$$

For reference purposes, we call the above model as *Nonparametric Hierarchical Factor Analysis* (NHFA) and the two variants as NHFA-PGM and NHFA-GGM.

## 4 Model Inference

For inference, we use a combination of slice sampling and collapsed Gibbs sampling. The main variables of interest for NHFA model are $\mathbf{Z}$, $\mathbf{W}$, $\Phi$, $\pi$, $\beta$ and $\alpha_j$ (see Figure 1). In addition, we introduce an auxiliary variable $\rho$. The purpose of this auxiliary variable is to turn the infinite representation of beta process into a finite representation. We integrate out $\pi$ and sample $\mathbf{Z}$, $\mathbf{W}$, $\Phi$ and $\beta$. Notation-wise, we use $\beta = \{\beta_k : k = 1, \ldots, K\}$, $\pi = \{\pi_j : j = 1, \ldots, J\}$ and a superscript attached to a variable followed by a '−' sign (e.g. $\mathbf{m}^{-jk}$, $\mathbf{Z}_j^{i,-k}$ etc) means a set excluding the variable indexed by the superscript. In the following, we describe the inference steps. ***Further details on some of the derivations are provided in the supplementary material [9].***

The joint distribution of $\beta$ and $\mathbf{Z}$ can be written as

$$p(\beta, \mathbf{Z}, \rho) = p(\beta) p(\rho \mid \mathbf{Z}, \beta) \int_\pi p(\mathbf{Z} \mid \pi) p(\pi \mid \beta) d\pi \quad (10)$$

Let us assume that the auxiliary variable $\rho$ is uniformly distributed as

$$p(\rho \mid \mathbf{Z}, \beta) = \frac{1}{\beta^*} \mathbf{1}(0 \leq \rho \leq \beta^*) \quad (11)$$

where $\beta^*$ (a function of $\beta_{(1:\infty)}$ and $\mathbf{Z}$) is equal to the stick weight of the smallest active factor, i.e.

$$\beta^* = \min_{k \mid \exists i, \, \mathbf{Z}_j^{ik} = 1} \beta_{(k)} \quad (12)$$

Using Eq (11), the conditional of $\mathbf{Z}$ given $\rho$ and $\beta$ can be written as

$$p(\mathbf{Z} \mid \rho, \beta) \propto \frac{1}{\beta^*} \mathbf{1}(0 \leq \rho \leq \beta^*) \int_\pi p(\mathbf{Z} \mid \pi) p(\pi \mid \beta) d\pi \quad (13)$$

We note that given $\rho$, $\forall i$, $\mathbf{Z}_j^{ik} = 0$ if $\beta_{(k)} < \rho$ and therefore, we only need to update $\mathbf{Z}_j^{ik}$ for those values of $k$ such that $\beta_{(k)} \geq \rho$ [17]. We now proceed to derive the full conditional distributions required for Gibbs sampling.

### 4.1 Sampling $\rho$

Let $K^\dagger$ denote the index such that all active features have index $k < K^\dagger$. The index $K^\dagger$ is also represented although it is an inactive feature. We sample $\rho$ according to Eq (11). If the new value of $\rho$ is such that $\beta_{(K^\dagger)} \geq \rho$, then we extend the stick breaking representation until $\beta_{(K^\dagger)} < \rho$. For the extended representation, $\Phi$ and $\mathbf{W}$ can be sampled from their prior distributions while $\beta_{(k)}$ are sampled from the following conditional distribution

$$p\left(\beta_{(k)} \mid \beta_{(k-1)}, \mathbf{Z}^{:,\geq k} = 0\right)$$
$$\propto p\left(\beta_{(k)} \mid \beta_{(k-1)}\right) p\left(\mathbf{Z}^{:,\geq k} = 0 \mid \beta_{(k)}\right) \quad (14)$$

where $\mathbf{Z}^{:,\geq k} \triangleq \left\{\mathbf{Z}_j^{ik'} \mid \forall j, i \text{ and } k' \geq k\right\}$. The likelihood term in the above expression can be computed as

$$p\left(\mathbf{Z}^{:,\geq k} = 0 \mid \beta_{(k)}\right)$$
$$= \Pi_{j=1}^J p\left(\mathbf{Z}_j^{:,k} = 0 \mid \beta_{(k)}\right) p\left(\mathbf{Z}_j^{:,>k} = 0 \mid \beta_{(k)}\right) \quad (15)$$

In the above expression, the first term of the right hand side (R.H.S.) requires marginalizing out $\pi_{jk}$ and is computed as

$$p\left(\mathbf{Z}_j^{:,k} = 0 \mid \beta_{(k)}\right)$$
$$= \int_0^1 p\left(\mathbf{Z}_j^{:,k} = 0 \mid \pi_{jk}\right) p\left(\pi_{jk} \mid \beta_{(k)}\right) d\pi_{jk}$$
$$= \frac{\Gamma(\alpha_j) \Gamma(\alpha_j \bar{\beta}_{(k)} + N_j)}{\Gamma(\alpha_j \bar{\beta}_{(k)}) \Gamma(\alpha_j + N_j)} \quad (16)$$

where $\bar{\beta}_{(k)} \triangleq 1 - \beta_{(k)}$. For each $k \geq 1$, let us use $l_k$ to define the mapping $\beta_{l_k} = \beta_{(k)}$ and let $\mathbf{L}_k = \{1, \ldots, K\} \setminus \{l_1, \ldots, l_k\}$, then the second term in the right hand side (R.H.S.) of Eq (15) can be computed as the following

$$p\left(\mathbf{Z}_j^{:,>k} = 0 \mid \beta_{(k)}\right)$$
$$= \lim_{K \to \infty} \int p\left(\beta_{\mathbf{L}_k} \mid \beta_{(k)}\right) p\left(\mathbf{Z}_j^{:,\mathbf{L}_k} = 0 \mid \beta_{\mathbf{L}_k}\right) d\beta_{\mathbf{L}_k}$$
$$= \lim_{K \to \infty} \int \Pi_{l \in \mathbf{L}_k} p\left(\beta_l \mid \beta_{(k)}\right) p\left(\mathbf{Z}_j^{:,l} = 0 \mid \beta_l\right) d\beta_{\mathbf{L}_k}$$
$$= \lim_{K \to \infty} \left(\int_0^{\beta_{(k)}} p\left(\beta_l \mid \beta_{(k)}\right) p\left(\mathbf{Z}_j^{:,l} = 0 \mid \beta_l\right) d\beta_l\right)^{K-k}$$
$$\quad (17)$$

In the above limmand, we first simplify the integral to the following

$$\int_0^{\beta_{(k)}} p\left(\beta_l \mid \beta_{(k)}\right) p\left(\mathbf{Z}_j^{:,l} = 0 \mid \beta_l\right) d\beta_l$$
$$= \frac{\Gamma(\alpha_j)}{\Gamma(\alpha_j + N_j)} \sum_{u_{jk}=0}^{N_j} \begin{bmatrix} N_j \\ u_{jk} \end{bmatrix} \alpha_j^{u_{jk}}$$
$$\times \frac{\Pi_{t''=1}^{u_{jk}} t''}{\Pi_{t'=1}^{u_{jk}} \left(\frac{\tau}{K} + t'\right)} \left[1 + \frac{\tau}{K} \sum_{p=1}^{u_{jk}} \frac{\Pi_{t=1}^{p-1}\left(\frac{\tau}{K} + t\right)}{p!} \bar{\beta}_{(k)}^p\right] \quad (18)$$

where $\begin{bmatrix} n \\ m \end{bmatrix}$ denotes the unsigned Stirling numbers of the first kind. In the above expression, we note that the term $\frac{\Gamma(\alpha_j)}{\Gamma(\alpha_j+N_j)} \sum_{u_{jk}=0}^{N_j} \begin{bmatrix} N_j \\ u_{jk} \end{bmatrix} \alpha_j^{u_{jk}} \times (\ldots)$ can be written as

$$\frac{\Gamma(\alpha_j)}{\Gamma(\alpha_j+N_j)} \sum_{u_{jk}=0}^{N_j} \begin{bmatrix} N_j \\ u_{jk} \end{bmatrix} \alpha_j^{u_{jk}} \times (\text{ term involving } u_{jk})$$
$$= \sum_{u_{jk}=0}^{N_j} \frac{\begin{bmatrix} N_j \\ u_{jk} \end{bmatrix} \alpha_j^{u_{jk}}}{\sum_{u'_{jk}=0}^{N_j} \begin{bmatrix} N_j \\ u'_{jk} \end{bmatrix} \alpha_j^{u'_{jk}}} \times (\text{ term involving } u_{jk})$$
$$= \sum_{u_{jk}=0}^{N_j} \tilde{\omega}_{u_{jk}} \times (\text{ term involving } u_{jk})$$

where $\sum_{u_{jk}=0}^{N_j} \tilde{\omega}_{u_{jk}}$ is equal to 1. Continuing from Eq (18) and substituting it in Eq (17), we can write

$$p\left(\mathbf{Z}_j^{:,>k} = 0 \mid \beta_{(k)}\right)$$
$$= \lim_{K \to \infty} \left(\sum_{u_{jk}=0}^{N_j} \tilde{\omega}_{u_{jk}} \frac{\Pi_{t''=1}^{u_{jk}} t''}{\Pi_{t'=1}^{u_{jk}} \left(\frac{\tau}{K} + t'\right)} \left(1 + \frac{\tau}{K} R_{u_{jk}}\right)\right)^{K-k}$$

which simplifies to (see *supplementary material [9]*)

$$p\left(\mathbf{Z}_j^{:,>k} = 0 \mid \beta_{(k)}\right)$$

$$= \lim_{K \to \infty} \left(1 + \sum_{u_{jk}=0}^{N_j} \tilde{\omega}_{u_{jk}} \frac{\tau \left(T_{u_{jk}} - H_{u_{jk}}\right)}{K + \tau H_{u_{jk}}}\right)^{K-k}$$

$$= \exp\left(\tau \sum_{u_{jk}=0}^{N_j} \tilde{\omega}_{u_{jk}} \left(T_{u_{jk}} - H_{u_{jk}}\right)\right) \quad (19)$$

where we have defined $R_{u_{jk}} \triangleq \sum_{p=1}^{u_{jk}} \frac{\prod_{t=1}^{p-1}\left(\frac{\tau}{K}+t\right)}{p!} \bar{\beta}_{(k)}^p$, $T_{u_{jk}} \triangleq \sum_{p=1}^{u_{jk}} \frac{\bar{\beta}_{(k)}^p}{p}$ and the harmonic number $H_{u_{jk}} = \sum_{h=1}^{u_{jk}} \frac{1}{h}$. Plugging the likelihood of (19) and (16) into Eq (15) and using Eq (14), we obtain

$$p\left(\beta_{(k)} \mid \beta_{(k-1)}, \mathbf{Z}^{:,\geq k} = 0\right)$$

$$\propto \tau \beta_{(k-1)}^{-\tau} \beta_{(k)}^{\tau-1} \mathbf{1}\left(0 \leq \beta_{(k)} \leq \beta_{(k-1)}\right) \times$$

$$\Pi_{j=1}^{J} \frac{\Gamma\left(\alpha_j \bar{\beta}_{(k)} + N_j\right)}{\Gamma\left(\alpha_j \bar{\beta}_{(k)}\right)} \exp\left(\tau \sum_{u_{jk}=0}^{N_j} \tilde{\omega}_{u_{jk}} \left(T_{u_{jk}} - H_{u_{jk}}\right)\right)$$

$$\propto \beta_{(k)}^{\tau-1} \mathbf{1}\left(0 \leq \beta_{(k)} \leq \beta_{(k-1)}\right) \times$$

$$\Pi_{j=1}^{J} \sum_{v_{jk}=0}^{N_j} \begin{bmatrix} N_j \\ v_{jk} \end{bmatrix} \left(\alpha_j \bar{\beta}_{(k)}\right)^{v_{jk}} \exp\left(\tau \sum_{u_{jk}=0}^{N_j} \tilde{\omega}_{u_{jk}} \sum_{p=1}^{u_{jk}} \frac{\bar{\beta}_{(k)}^p}{p}\right)$$

Introducing the auxiliary variables $\mathbf{v} = \left(v_{jk} : \forall j, k\right)$, we get

$$p\left(\beta_{(k)} \mid \beta_{(k-1)}, \mathbf{Z}^{:,\geq k} = 0, \mathbf{v}\right)$$

$$\propto \beta_{(k)}^{\tau-1} \left(\bar{\beta}_{(k)}\right)^{\sum_j v_{jk}} \exp\left(\tau \sum_{u_{jk}=0}^{N_j} \tilde{\omega}_{u_{jk}} \sum_{p=1}^{u_{jk}} \frac{\bar{\beta}_{(k)}^p}{p}\right)$$

$$\mathbf{1}\left(0 \leq \beta_{(k)} \leq \beta_{(k-1)}\right) \quad (20)$$

The auxiliary variable $v_{jk}$ can be sampled as

$$p\left(v_{jk} \mid \mathbf{v}^{-jk}\right) \propto \begin{bmatrix} N_j \\ v_{jk} \end{bmatrix} \left(\alpha_j \bar{\beta}_{(k)}\right)^{v_{jk}} \quad (21)$$

Sampling from Eq (20) can be obtained using Adaptive Rejection Sampling (ARS) procedure as the distribution is *log-concave* in $\beta_{(k)}$. Sampling of auxiliary variable $v_{jk}$ is straight-forward as it is sampling from a discrete distribution with finite support.

### 4.2 Sampling $\mathbf{Z}_j$

Given other variables, $\mathbf{Z}_j^{ik}$ (for $k = 1, \ldots, K^\dagger - 1$) can be sampled from the following Gibbs conditional posterior distribution

$$p\left(\mathbf{Z}_j^{ik} = 1 \mid \mathbf{Z}_j^{i,-k}, \mathbf{Z}^{-ji}, \mathbf{W}_j^i, \beta, \Phi, \mathbf{X}_j^i, \rho\right)$$

$$\propto p\left(\rho \mid \mathbf{Z}_j^{ik} = 1, \mathbf{Z}_j^{i,-k}, \mathbf{Z}^{-ji}, \beta\right)$$

$$\times p\left(\mathbf{Z}_j^{ik} = 1 \mid \mathbf{Z}_j^{-i,k}, \beta_{(k)}\right) p\left(\mathbf{X}_j^i \mid \mathbf{Z}_j^{ik} = 1, \mathbf{Z}_j^{i,-k}, \mathbf{W}_j^i, \Phi\right)$$

$$= \frac{n_j^{-i,k} + \alpha_j \beta_{(k)}}{\beta^*} p\left(\mathbf{X}_j^i \mid \mathbf{Z}_j^{ik} = 1, \mathbf{Z}_j^{i,-k}, \mathbf{W}_j^i, \Phi\right) \quad (22)$$

where $n_j^{-i,k} = \sum_{i' \neq i} \mathbf{Z}_j^{i'k}$. In the above posterior expression, we note the dependency of $\beta^*$ on the sample of $\mathbf{Z}_j^{ik}$.

### 4.3 Sampling $\beta_{(k)}$

To sample from the posterior of each $\beta_{(k)}$, we use auxiliary variable sampling [3, 18]. We define $\mathbf{m} = \left(m_{jk} : \forall j, k\right)$, $\mathbf{l} = \left(l_{jk} : \forall j, k\right)$ and $n_{jk} = \sum_i \mathbf{Z}_j^{i,k}$ where $m_{jk} \in \{0, 1, \ldots, n_{jk}\}$, $l_{jk} \in \{0, 1, \ldots, N_j - n_{jk}\}$, and sample $\beta_{(k)}$ (for $k = 1, \ldots, K^\dagger$) and auxiliary variables $\mathbf{m}, \mathbf{l}$ as below

$$p\left(\beta_{(k)} \mid \beta^{-(k)}, \mathbf{m}, \mathbf{l}, \mathbf{Z}, \rho\right)$$

$$\propto \beta_{(K^\dagger)}^{\tau_0} \beta_{(k)}^{m_k - 1} \left(1 - \beta_{(k)}\right)^{l_k} \mathbf{1}\left(\beta_{(k+1)} \leq \beta_{(k)} \leq \beta_{(k-1)}\right)$$
(23)

$$p\left(m_{jk} \mid \mathbf{m}^{-jk}, \mathbf{l}, \beta, \mathbf{Z}, \rho\right) \propto \begin{bmatrix} n_{jk} \\ m_{jk} \end{bmatrix} \left(\alpha_j \beta_k\right)^{m_{jk}} \quad (24)$$

$$p\left(l_{jk} \mid \mathbf{l}^{-jk}, \mathbf{m}, \beta, \mathbf{Z}, \rho\right) \propto \begin{bmatrix} \bar{n}_{jk} \\ l_{jk} \end{bmatrix} \left(\alpha_j \bar{\beta}_k\right)^{l_{jk}} \quad (25)$$

where $\mathbf{m}_k \triangleq \sum_j m_{jk}$, $\mathbf{l}_k \triangleq \sum_j l_{jk}$, $\bar{n}_{jk} = N_j - n_{jk}$. For detailed derivation, see *supplementary material [9]*.

### 4.4 Sampling Parameters $\Phi$, $\mathbf{W}_j$ and $\lambda_j$ for NHFA-PGM model

Under the model described in Eq (8), Gibbs conditional posterior of $\Phi$ can be written as

$$p\left(\Phi^{i,:} \mid \mathbf{Z}_{1:J}, \mathbf{W}_{1:J}, \mathbf{X}_{1:J}, \lambda_{1:J}, a_\phi, b_\phi, \mathbf{s}\right)$$

$$\propto \Pi_{k=1}^{K^\dagger} \left(\Phi^{ik}\right)^{a_\phi + \sum_{j,l} s_j^{ilk} - 1} \exp\left\{-\left(b_\phi + \sum_{j,l} \mathbf{H}_j^{lk}\right) \Phi^{ik}\right\}$$
(26)

The auxiliary variables $\mathbf{s} = \left\{s_j^{ilk}, \forall j, l\right\}_{k=1}^{K^\dagger + 1}$ are drawn as

$$p\left(s_j^{il1}, \ldots s_j^{ilK^\dagger}, s_j^{il(K^\dagger + 1)} \mid \text{rest}\right)$$

$$\propto \frac{\mathbf{X}_j^{il}!}{s_j^{il1}! \ldots s_j^{ilK^\dagger}! s_j^{il(K^\dagger + 1)}!} \Pi_{k=1}^{K^\dagger} \left(\Phi^{ik} \mathbf{H}_j^{lk}\right)^{s_j^{ilk}} \lambda_j^{s_j^{il(K^\dagger + 1)}}$$

$$\text{s.t. } \sum_{p=1}^{K^\dagger + 1} s_j^{ilp} = 1$$
(27)

Gibbs sampling update for $\mathbf{W}_{ji}$ conditioned on the remaining variables is given as

$$p\left(\mathbf{W}_j^{l,:} \mid \mathbf{Z}_j, \Phi, \mathbf{X}_j, \lambda_j, a_{w_j}, b_{w_j}, \mathbf{t}\right)$$

$$\propto \Pi_{k=1}^{K^\dagger} \left(\mathbf{W}_j^{lk}\right)^{a_{w_j} + \sum_i t_j^{ilk} - 1}$$

$$\times \exp\left\{-\left(b_{w_j} + \sum_i \left(\Phi \mathbf{D}_{\mathbf{Z}_j^{l,:}}\right)^{ik}\right) \mathbf{W}_j^{lk}\right\} \quad (28)$$

where $\mathbf{D}_{\mathbf{Z}_j^{l,:}}$ denotes a diagonal matrix constructed from $\mathbf{Z}_j^{l,:}$ and the auxiliary variables $\mathbf{t} = \left\{ t_j^{ilp}, \text{for each } i \right\}_{p=1}^{K^\dagger+1}$ are sampled as below

$$p\left(t_j^{il1}, \ldots t_j^{ilK^\dagger}, t_j^{il(K^\dagger+1)} \mid \text{rest}\right)$$
$$\propto \frac{\mathbf{X}_j^{il}!}{t_j^{il1}! \ldots t_j^{ilK^\dagger}! t_j^{il(K^\dagger+1)}!} \Pi_{k=1}^{K^\dagger} \left(\Phi^{lk} \mathbf{H}_j^{lk}\right)^{t_j^{ilk}} \lambda_j^{t_j^{il(K^\dagger+1)}}$$
$$\text{s.t. } \sum_{p=1}^{K^\dagger+1} t_j^{ilp} = 1 \qquad (29)$$

Gibbs sampling of $\lambda_j$ remains similar to the variables $\Phi$ and $\mathbf{W}_j$. The required posterior distributions are given as

$$p(\lambda_j \mid \mathbf{Z}_{1:J}, \mathbf{W}_{1:J}, \Phi, \mathbf{X}_{1:J}, \mathbf{r})$$
$$\propto \lambda_j^{a_{\lambda_j} + \sum_{i=1}^{M} \sum_{l=1}^{N_j} r_j^{il} - 1} \exp\left\{-\left(b_{\lambda_j} + MN_j\right)\right\} \qquad (30)$$

For each $i, l$, the auxiliary variables $r_j^{il}$ can be sampled as

$$r_j^{il} \mid \text{rest} = \text{Binomial}\left(\mathbf{X}_j^{il}, \frac{\lambda_j}{\Phi^{i,:}\left(\mathbf{H}_j^{l,:}\right)^\mathsf{T} + \lambda_j}\right) \qquad (31)$$

### 4.5 Sampling Parameters $\Phi$, $\mathbf{W}_j$, $\sigma_{w_j}$ and $\sigma_{n_j}$ for NHFA-GGM model

Gibbs sampling updates for these parameters remains same as the updates described in [8].

### 4.6 Sampling $\alpha_j$

Once again, Gibbs sampling updates for $\alpha_j$ remains same as the updates described in [8]. The hyperparameter $\alpha_j$ controls the variation of $A_j$ (source-specific beta process) around $B$ (parent beta process). As $\alpha_j$ vary from a low to high value, its concentration on the random distribution $B$ increases. Since the random distribution $B$ is shared across different data sources, this increases the probability of sharing more and more factors.

### 4.7 Predictive Likelihood

Let us denote the test data from the $j$-th source by matrix $\tilde{\mathbf{X}}_j$ and the corresponding matrix factorization as $\tilde{\mathbf{X}}_j = \Phi\left(\tilde{\mathbf{Z}}_j \odot \tilde{\mathbf{W}}_j\right) + \tilde{\mathbf{E}}_j$, then the Monte Carlo approximation of predictive likelihood can be written as

$$p\left(\tilde{\mathbf{X}}_j \mid \mathbf{X}_{1:J}\right) \approx \frac{1}{LR} \sum_{r=1}^{R} \sum_{l=1}^{L} p\left(\tilde{\mathbf{X}}_j \mid \Phi^{[l]}, \tilde{\mathbf{Z}}_j^{[r]}, \tilde{\mathbf{W}}_j^{[r]}\right) \qquad (32)$$

where $L$ is the number of training samples $\left\{\Phi^{[l]}\right\}$ and $R$ is the number of test samples $\left\{\tilde{\mathbf{Z}}_j^{[r]}, \tilde{\mathbf{W}}_j^{[r]}\right\}$.

## 5 Experiments

We carry out a variety of experiments to demonstrate the effectiveness of the proposed NHFA model. To illustrate the behavior of our model, we first perform experiments with a synthetic dataset, for which the true dimensionality of subspaces and other parameters are known. Through these experiments, we show the correct recovery of these parameters. Next, we demonstrate the usefulness of our model for two real-world tasks – text modeling and content based image retrieval. For both synthetic and real-world tasks, the priors for hyperparameters are chosen as the following : $\tau_0 = 1$, $\alpha_j \sim \text{gamma}(1,1)$. For NHFA-GGM, both the shape and the scale parameters of gamma priors for $\sigma_\phi$, $\sigma_{nj}$ and $\sigma_{wj}$ were set to 1. For NHFA-PGM, since we expect the results to be sparse we set the shape parameters of hyperparameters as $a_\phi = 1$, $a_{w_j} = 1$ whereas the scale parameters were sampled as : $b_\phi \sim \text{gamma}\left(1, 1/\mu_\phi\right)$, $b_{w_j} \sim \text{gamma}\left(1, 1/\mu_{w_j}\right)$, where $\mu_\phi \triangleq \frac{1}{MK} \sum_{i,k} \Phi^{ik}$ and $\mu_{w_j} \triangleq \frac{1}{N_j K} \sum_{l,k} \mathbf{W}_j^{lk}$. *Supplementary material* provides further details.

### 5.1 Experiments-I : Synthetic Data

We create a synthetic dataset similar to [8] so that we can show the benefits of our model vis-à-vis the model considered in [8]. For this dataset, we create twelve 100-D binary factors and distribute them across the two data sources termed as $\mathcal{D}_1$ and $\mathcal{D}_2$. The first four factors were used by the data points from $\mathcal{D}_1$, the next four factors were used by the data points from $\mathcal{D}_2$ while the last four factors were shared by data points across both $\mathcal{D}_1$ and $\mathcal{D}_2$. We generated the mixing configuration matrices $\mathbf{Z}_1$ and $\mathbf{Z}_2$ randomly with discrete support $\{0, 1\}$. The weight/coefficient matrices of the two sources, i.e. $\mathbf{W}_1$ and $\mathbf{W}_2$ were sampled from gamma distribution with shape and scale parameters 1 and 2 respectively. The using these parameters along with noise (with rate parameter 0.1 for both sources), the data for both $\mathcal{D}_1$ and $\mathcal{D}_2$ were generated from Poisson distributions according to the generative model described in section 3.

To verify the correctness of the inference, we run the slice sampler along with Gibbs updates as detailed in section 4 starting with the value of $K^\dagger$ as one. We observe that the sampler converges to the true value of the number of factors (i.e. $K^\dagger = 12$) in less than 100 iterations. However, we run the sampler longer to verify that the mode of the posterior over the number of factors remains at this true value. It can be seen from Figure 2 that the model correctly learns all the factors and their scores automatically.

To compare the computational efficiency of our slice sampler with the approximate Gibbs sampler of [8], we see that both methods need to sample $\{\beta, \mathbf{Z}, \Phi, \mathbf{W}\}$. Sampling $\Phi$ and $\mathbf{W}$ remain identical in both cases. Only sampling of $\beta$ and $\mathbf{Z}$ differ. In case of the approximate Gibbs sampler, $\beta$ conditioned on $\{\mathbf{m}, \mathbf{l}, \mathbf{Z}\}$ is drawn from a beta distribution whereas in the slice sampler, $\beta$ conditioned on $\{\rho, \mathbf{m}, \mathbf{l}, \mathbf{Z}\}$ uses adaptive rejection sampling (ARS) (see Eq (23)). Although not as fast as sampling from a beta distribution, ARS is quite efficient due to sampling in one dimensional space. When comparing sampling of $\mathbf{Z}$, the slice sampler is more efficient than the approximate Gibbs sampler. The main difference lies in sampling the number of new factors for $i$-th data point from $j$-th source (say $F_{ji}$). In slice sampling, $F_{ji}$ grows gradually based on the slice variable and requires sampling of $F_{ji}$ new $\beta_k$ variables using ARS. On the other hand, the approximate Gibbs sampler, to sample $F_{ji}$, approximates an intractable integral (Eq (3.22) in [8]) using Monte Carlo samples. For the synthetic experiments described above, slice sampler takes about *100 iterations* to converge and runs in **6.67** minutes. On the other hand, the approximate Gibbs sampler [8] requires about *1000 iterations* to converge to the true number of factors taking totally **52.3** minutes. This timing analysis is performed using a Windows PC with Intel i7@3.4 GHz and 8 GB RAM.

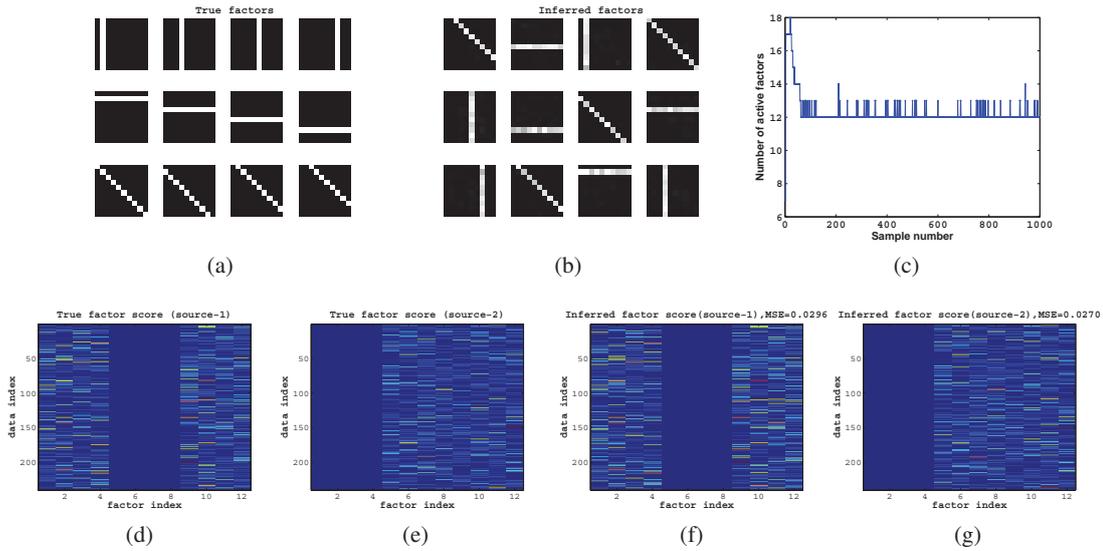

Figure 2: Synthetic data experiments (a) *true* factors (b) *inferred* factors (c) Number of active factors vs. Gibbs iterations (d) *true* factor scores for source-1 (e) *true* factor scores for source-2 (f) *inferred* factor scores for source-1 (g) *inferred* factor scores for source-2; for an easy comparison with (d) and (e), columns of (f) and (g) are permuted according to the mapping between (a) and (b).

### 5.2 Experiments-II : Real Data

#### 5.2.1 Results using NIPS 0-12 Dataset

Our first real-world dataset is the NIPS 0-12 dataset, which contains the articles from the proceedings of *Neural Information Processing Systems* (NIPS) conference published between 1988 and 1999. In this dataset[1], text articles are divided into nine different sections/tracks plus one miscellaneous section/track. We work with nine sections which are Cognitive Science (CS), Neuroscience (NS), Learning Theory (LT), Algorithms and Architecture (AA), Implementations (IM), Speech and Signal Processing (SSP), Visual Processing (VP), Applications (AP) and Control, Navigation and Planning (CNP). We treat each section as one data source and generate nine different term-document matrices. In doing so, we ensure that these matrices use a common dictionary for terms[2].

---
[1] A processed version of this dataset is made available by Y. W. Teh at http://www.gatsby.ucl.ac.uk/~ywteh/research/data.html.
[2] It is possible to make a common dictionary by merging the terms from all the sections.

Through this dataset, we intend to show the strength of our model for transfer learning. For this, we choose section VP as target data source while other sections (one at a time) as auxiliary data sources. We follow this scheme for *two* reasons. *First*, we believe that although there may be underlying sharing across different sections, each section has its own focus and differs in distribution. NHFA exploits the sharing across different sections while still retaining the focus of individual section by maintaining a hierarchy. *Second*, this allows us to compare our results with two baselines (i.e. Gupta et al [8] and Teh et al [18]) as they use the same dataset with similar settings.

Out of 1564 articles in total across nine sections, we randomly select 80 articles from each section and use them for training. Similar to [8, 18], the test set is chosen from VP section (consisting of 44 articles) and kept fixed throughout the experimentation with NIPS 0-12 dataset. On average, the number of words per article are approximately 1000. We compute the perplexity on the test set and report the performance in terms of *perplexity per document*. Given the training data $\mathbf{X}_{1:J}$ and a test set $\tilde{\mathbf{X}}_j$ from $j$-th data source, perplexity per document (PPD) is defined as

$$\text{PPD}\left(\tilde{\mathbf{X}}_j\right) = \exp\left(-\frac{1}{\tilde{N}}\log p\left(\tilde{\mathbf{X}}_j \mid \mathbf{X}_{1:J}\right)\right) \quad (33)$$

where $\tilde{N}$ denotes the number of documents in the test set.

**Baseline Methods** To compare the proposed model with other related works, we choose *three* baselines.

- Baseline-1a ["No auxiliary"]: This baseline is a linear Poisson-gamma based factor analysis model which totally relies on the target data and does not use any auxiliary data for training.

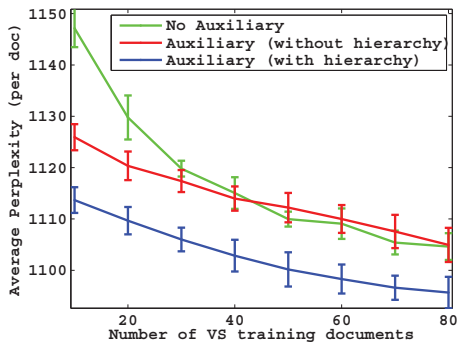 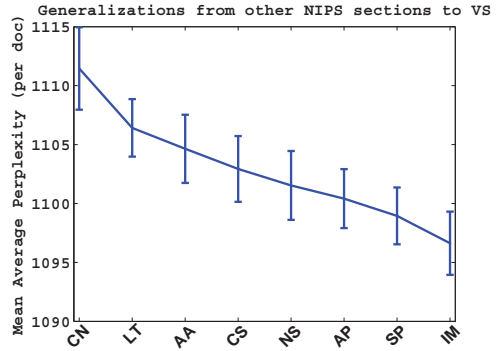

(a)             (b)

Figure 3: Perplexity results using NIPS 0-12 dataset (a) perplexity for the test data from VP section vs. number of VP training documents (averaged over 10 runs); for Baseline-1a, Baseline-1b and the proposed NHFA-PGM (b) mean average perplexity for the test data from VP section (averaged over 10 runs) versus different NIPS auxiliary sections (except VP) using the proposed NHFA-PGM.

- Baseline-1b ["Auxiliary - without hierarchy"]: This baseline is a linear Poisson-gamma based factor analysis model similar to the first baseline but also uses auxiliary data simply by augmenting them with the target data and does not maintain any hierarchy.

- Baseline-2 [NHFA-GGM (approx.)]: This baseline is the hierarchical model recently proposed in Gupta et al [8]. This model attempts to learn both shared and individual subspace using a hierarchical model. The data distributions for this model are similar to the NHFA-GGM model proposed in section 2, however it uses a series of approximations for inference.

- Baseline-3 [HDP]: This baseline is the hierarchical Dirichlet process proposed in [18]. This model uses a hierarchy for sharing topics across different groups.

**Experimental Results** We run the proposed NHFA-PGM and the baseline models and compute the perplexity per doc (PPD) values over 10 independent runs. For each run, we select VP section as the target source while other sections as auxiliary (one at a time) and compute perplexity values for increasing number of target training articles varying from 10 to 80 at a step of 10. By doing this, our intention is to see at what point, the auxiliary data ceases to improve the generalization performance.

Figure 3 depicts the perplexity results for the proposed model and compares with baseline-1a and baseline-1b. It can be seen from Figure 3a that the proposed NHFA-PGM model achieves significantly lower perplexity compared to the two baselines irrespective of the number of training articles from VP section. When comparing the results between the baseline-1a and the baseline 1b, one can see that as long as the number of training documents from VP section are less than 50, the performance of baseline-1b is better than that of baseline 1a. This goes according to our intuition as the auxiliary source help learning some patterns.

But, as soon as there are enough training articles from the VP section, simply combining auxiliary data does not help and the performance of baseline-1b goes lower than that of baseline-1a. This clearly shows that the baseline-1b is prone to negative transfer learning whereas the same is not true for the proposed NHFA-PGM model which retains low perplexity values compared to baseline-1a. Figure 3b focuses on the results of NHFA-PGM model and provides the mean average perplexity for each auxiliary section (a single figure measure computed by averaging the perplexity values along increasing number of training documents). It can be seen from the figure that the top three auxiliary sections that help maximum generalization are IM, SP and AP in decreasing order of performance.

Table 1 lists a comparison of the proposed NHFA-PGM and NHFA-GGM models with baseline-2 and baseline-3. One can clearly see that since the NIPS 0-12 data matrices are count data, both NHFA-PGM and the baseline-3 achieve significantly lower perplexity compared to NHFA-GGM and baseline-2. The proposed NHFA-PGM clearly outperforms the baseline-3, suggesting that Poisson-gamma model may be a better fit to the count data than topic models. When comparing the perplexity values of baseline-2 and NHFA-GGM, NHFA-GGM achieves better performance. This gain in performance is attributed to the exact inference made possible by slice sampling.

| Method | Average Log Perplexity (per doc) |
|---|---|
| NHFA-GGM (approx.) [8] | $6232.5 \pm 164.0$ |
| Proposed NHFA-GGM | $5945.3 \pm 16.8$ |
| HDP [18] | $2862.5 \pm 268.3$ |
| **Proposed NHFA-PGM** | **$1102.9 \pm 6.5$** |

Table 1: Comparison of various hierarchical models in terms of average log perplexity per doc (PPD) using NIPS 0-12 dataset.

### 5.2.2 Results using NUS-WIDE Animal Dataset

Our second dataset is a subset of the NUS-WIDE [2] dataset. This subset comprises of 3411 images involving 13 animals. We use six different low-level features [2] namely 64-D color histogram, 144-D color correlogram, 73-D edge direction histogram, 128-D wavelet texture, 225-D block-wise color moments. We construct a real-valued feature matrix for each animal category and treat it as a separate data source under our hierarchical framework. This is done with a belief that features of different animal categories vary in their distributions. Out of 3411 images, 2054 images are used for training while the remaining images are held for testing. This is done for comparing with [21, 22, 1, 8] where an *identical* training and test settings were used.

Unlike NIPS 0-12 data, the feature matrices for NUS-WIDE images are real valued. Therefore, we use NHFA-GGM model (instead of NHFA-PGM model) to learn the factor matrix $\Phi$ and $\mathbf{H}_j$ (for $j = 1, \ldots, 13$). Having learnt the matrix $\Phi$, we construct the test matrix $\tilde{\mathbf{X}}$ from the test data and compute its subspace coefficient matrix $\tilde{\mathbf{H}} = \left[\tilde{\mathbf{h}}_1, \ldots, \tilde{\mathbf{h}}_{\tilde{N}}\right]$ such that $\tilde{\mathbf{X}} \approx \Phi\tilde{\mathbf{H}}$. To retrieve the similar images for the $r$-th test image, cosine similarity between its subspace coefficient vector $\tilde{\mathbf{h}}_r$ and the subspace coefficient matrices of the training data (i.e. $\mathbf{H}_j$ for each $j = 1, \ldots, 13$) is computed. The retrieved images are ranked in decreasing order of their similarity values.

**Evaluation Measures and Baselines** To evaluate the performance of the proposed method, we use standard precision-scope curve[3] and mean average precision (MAP). We compare the result of the proposed NHFA-GGM with the recent state-of-the-art multi-view learning techniques [1, 21, 22] and "NHFA-GGM (approx.)" model [8] (the model used as baseline-2 for NIPS 0-12 experiments).

**Experimental Results** Table 2 compares the proposed NHFA-GGM with the recent works [1, 21, 22, 8] on image retrieval using NUS-WIDE animal dataset. This comparison is based on the mean average precision (MAP) values presented in [1, 8]. We note that the dataset used for generating these results (including the test set) is *identical*. For the works in [1, 21, 22], the MAP values are reported using 60 topics. The NHFA-GGM (approx.) model of [8] is a nonparametric model and reports the total number of shared and individual factors varying between 5 to 8. On the contrary, our proposed NHFA-GGM, which is exact version of NHFA-GGM (approx.), uses 30 to 40 factors typically. It can be clearly seen from the Table 2, that NHFA-GGM outperforms all the baselines. Focusing on the exact and approximate models, we report the precision scope curve which shows that NHFA-GGM model achieves clearly higher precision at initial scope levels which is often quite important in search applications as users are typically more interested in top few search results.

| Method | Mean Average Precision (MAP) |
|---|---|
| DWH [21] | 0.153 |
| TWH [22] | 0.158 |
| MMH [1] | 0.163 |
| NHFA-GGM (approx.) [8] | $0.1789 \pm 0.0128$ |
| **Proposed NHFA-GGM** | **$0.1945 \pm 0.0115$** |

Table 2: Comparison of image retrieval results with recent state-of-the art multi-view learning and hierarchical modeling techniques using NUS-WIDE animal dataset.

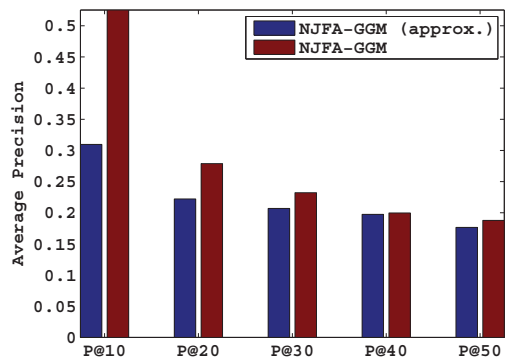

Figure 4: Comparison of the retrieval performances in terms of precision-scope (P@N) curve for NHFA-GGM and NHFA-GGM (approx.)[8].

## 6 Conclusion

We have presented a novel slice sampler for restricted hierarchical beta process (R-HBP). A key feature of the proposed sampler is that it keeps the inference tractable, even when the prior and the likelihood are non-conjugate – therefore, it can be applied for joint modeling of multiple data sources in more elaborate settings. Another key feature of this sampler is that it offers an exact inference and does not need a series of approximations used by standard Gibbs samplers [11, 16, 8]. We have combined this sampler with two hierarchical factor analysis data models – linear Poisson-gamma model and linear Gaussian-Gaussian model for modeling count data and real-valued data respectively. To show the utility of the proposed models, we applied them for learning shared and individual subspaces across multiple data sources where unlike parametric models, the subspace dimensionalities were learned automatically. Using two real world datasets (NIPS 0-12 and NUS-WIDE animal datasets), we have demonstrated that our method outperforms recent state-of-the-art methods for text modeling and content based image retrieval. Further, the slice sampling derived in this paper is general and can be utilized for other matrix factorizations.

---
[3]Precision scope (P@N) curve reports precision values for top N retrieved items.